\documentclass[10pt,twocolumn,letterpaper]{article}

\usepackage{iccv}
\usepackage{times}
\usepackage{epsfig}
\usepackage{graphicx}
\usepackage{amsmath}
\usepackage{amssymb}
\usepackage{tabularx}
\usepackage{multicol}
\usepackage{float}
\usepackage{amsmath}
\usepackage{makecell}
\DeclareMathOperator*{\argmax}{arg\,max}

\usepackage[breaklinks=true,bookmarks=false]{hyperref}

\iccvfinalcopy 

\def\iccvPaperID{****} 

\ificcvfinal\fi

\begin{document}

\title{Semantic Parsing of Colonoscopy Videos with Multi-Label Temporal Networks}

\author{Ori Kelner\thanks{Equal Contribution} \ \ \ \ \ Or Weinstein$^\ast$ \ \ \ \ \ Ehud Rivlin \ \ \ \ \ Roman Goldenberg \\ 
Verily Life Sciences\\
{\tt\small \{orikelner, orw, ehud, rgoldenberg\}@google.com}
}

\maketitle
\ificcvfinal\thispagestyle{empty}\fi

\begin{abstract}
Following the successful debut of polyp detection and characterization, more advanced automation tools are being developed for colonoscopy. The new automation tasks, such as quality metrics or report generation, require understanding of the procedure flow that includes activities, events, anatomical landmarks, etc. In this work we present a method for automatic semantic parsing of colonoscopy videos. The method uses a novel DL multi-label temporal segmentation model trained in supervised and unsupervised regimes. We evaluate the accuracy of the method on a test set of over 300 annotated colonoscopy videos, and use ablation to explore the relative importance of various method's components.
\end{abstract}

\section{Introduction}

\begin{figure}[t]
\vspace{-2ex}
    \centering
    \includegraphics[height=5cm]{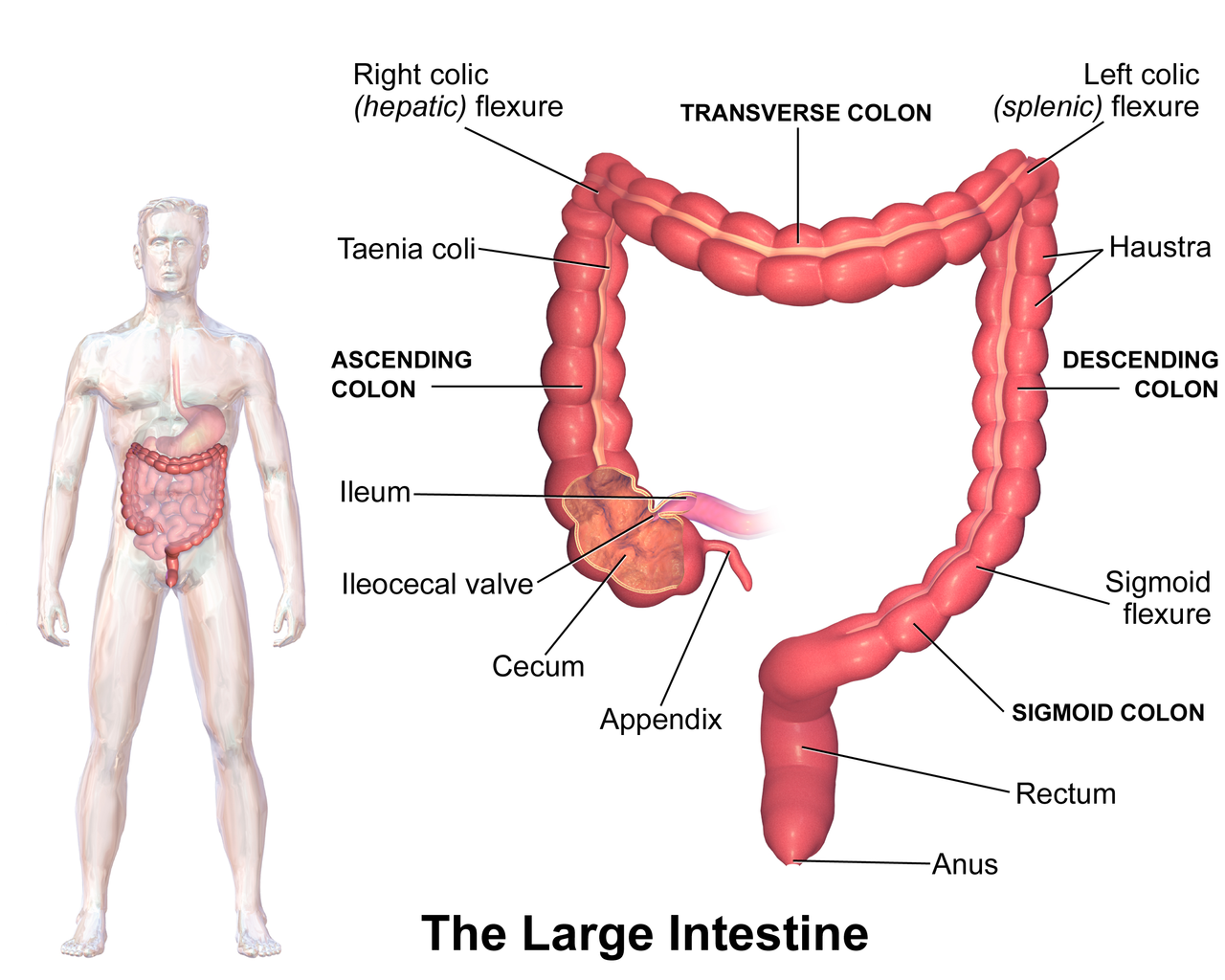} 
    \caption{Colon Anatomy (from 
    \href{https://upload.wikimedia.org/wikipedia/commons/2/2d/Blausen_0604_LargeIntestine2.png}{\texttt{Wikipedia}})}
    \label{fig:colon_map}
\vspace{-2ex}
\end{figure}

Optical colonoscopy is the standard of care procedure for colorectal cancer (CRC) screening. The primary target of screening colonoscopy is detecting polyps and preventively removing them. The first phase of the procedure (intubation) is inserting the endoscope all the way to the end the colon (cecum - see Fig.~\ref{fig:colon_map}). This is followed by the second phase (withdrawal), when the endoscope is slowly pulled out, while examining the colon mucosa for the presence of lesions. For some symptomatic indications, it is recommended to go farther than the cecum, into the terminal ileum, which is the final part of the small intestine. When examining the rectum, it is recommended to deflect the endoscope camera backwards in a U-turn (rectal retroflextion maneuver) to allow better visualization of the distal rectum. When a polyp is detected, it is often resected or biopsied using tools inserted through the colonoscope instrument channel. In some cases, the colonoscope can be taken out of the body and re-inserted during the procedure. 

The quality of the colonoscopy procedure is highly operator dependent and depends on the physician's skills, experience, fatigue, etc. To ensure high quality levels, professional societies recommend measuring and monitoring various quality metrics. For example, since not every procedure completes all recommended steps, cecal and ileum intubation rates and rectal retroflexion rate (percentage of procedures) are measured.

While the computer-aided tools for colonoscopy have been developed for years, only recently the first such tool - a polyp computed-aided detector (CADe) \cite{danny,detector,LACHTER2023,ou2021polyp,pacal2021robust}, became commercially available~\cite{detector,brand2022frame}. This success triggered the development of other, more advanced computer-aided tools for colonoscopy, including automatic quality metrics, colonoscopy video annotation and retrieval, automatic report generation~\cite{detector,survey,content_based}. A common prerequisite for those tasks is the ability to parse a colonoscopy video into semantically meaningful parts, including activities/phases (e.g. intubation, withdrawal, polyp management, cleansing), events/key moments (e.g. polyp detected, retroflextion), anatomical landmarks (e.g. ileocecal valve) and segments (e.g. rectum, cecum, inside/outside the body, etc.).

The ability to automatically parse procedures has a lot of potential to improve current practice. Below we present some use-cases:
\begin{itemize}
    \item Cecum detection enables straightforward calculation of withdrawal time \cite{liran_withdrawal}, which is a standard quality metric that estimates the amount of time the physician is looking for polyps.
    \item Combined with polyp detection capabilities, the proposed method would enable localization of polyps within the colon. This is crucial as some segments are more likely to contain precancerous polyps (adenomas).
    \item As mentioned earlier, semantic video parsing is a first step towards automatic report generation. This ability would potentially save time for physicians and could increase procedures volume.
    \item Detecting outside-body/inside-body allows removing the outside of the body video segments, which is a privacy requirement for using the colonoscopy videos for various research and clinical purposes.
    \item  The ability to detect tools allows computing the so-called "net withdrawal" time, which is the withdrawal time minus the time spent on polyp management. Net withdrawal time is a novel quality metric that might be better correlated to important clinical metrics such as adenoma detection rate (ADR).
\end{itemize}

Previous works on colonoscopy parsing were evaluated on significantly smaller datasets of around 20 videos \cite{parsing,parsing_2,parsing_3}. \cite{parsing_3}, for example, uses a boundary detection algorithm that detects changes in blurriness and pixel intensity between the segments. 
A number of works \cite{liran_withdrawal,unsupervised_withdrawal,withdrawal_3} focus on parsing colonoscopy into withdrawal and intubation phases. Similar in spirit to our work are \cite{lap_chole,lap_chole_2} that perform temporal segmentation of laparoscopic videos into surgical phases. Specifically, \cite{lap_chole} uses temporal networks on top of per frame features to detect surgical phases.    
 
\begin{figure*}[t]
    \centering
    \includegraphics[width=0.8\textwidth]{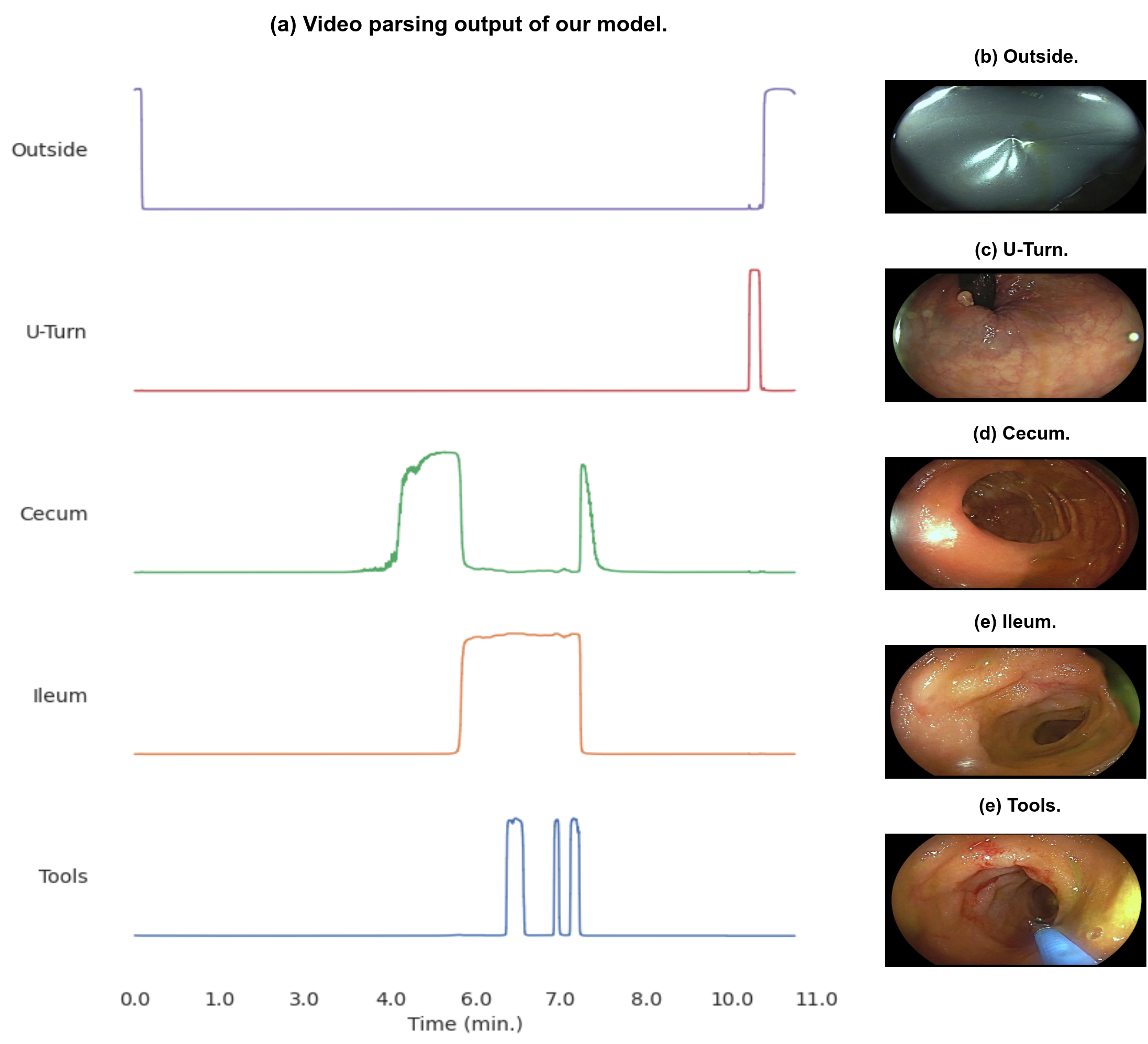} 
    \caption{The output of our model: class probabilities over the course of the procedure and the corresponding video snapshots.}
    \label{fig:colon_segments_2}
\end{figure*}

In what follows we present a method for colonoscopy video parsing and automatic detection of cecum, ileum, frames inside and outside of the body, rectal retroflexion, and use of surgical tools (see Fig. \ref{fig:colon_segments_2}).

The main paper contributions are:
\begin{itemize}
    \item A method to parse colonoscopy videos achieving 94.6\% balanced accuracy on a large test set of 344 videos. 
    \item An adaptation of a temporal convolution network to support multiple labels.
    \item A pseudo-labeling approach to increase the training set.
\end{itemize}


\section{Methods}

\begin{figure}[t]
    \vspace{-2ex}
    \centering
    \includegraphics[height=4.8cm]{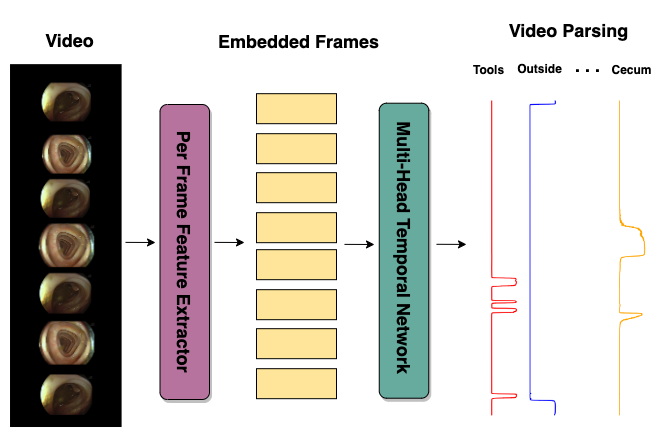} 
    \caption{The two stage video parsing pipeline. The first stage is a single frame encoder. The second stage runs temporal convolution (MS-TCN, ASFormer) on frame embeddings to yield per-frame classifications.}
    \label{fig:inference_pipeline}
    \vspace{-2ex}
\end{figure}

We follow a widely used two stage paradigm~\cite{mstcn} for video parsing: 
first, extracting features from video frames using a single frame encoder, and then feeding them into a temporal classifier that captures high-level temporal patterns (see Fig. \ref{fig:inference_pipeline}).

We propose several improvements to this straightforward approach, some of which are applicable to a wide range of scenarios and tasks, and some use the specific colonoscopy domain knowledge. 

\subsection{Baseline Method}
Our data consists of colonoscopy videos, annotated with the target labels (ileum, cecum, outside, inside, tools and rectal retroflextion) in the form of video segments, i.e. [start frame, end frame, label] triplets (see Section~\ref{experiments} for detailed dataset description).
The straightforward approach for video parsing is to start with a pretrained (e.g. on ImageNet~\cite{imagenet}) CNN, and use it as a single frame feature extractor for a temporal model. The temporal model, e.g. Multi-Stage Temporal Convolutional Network (MS-TCN) \cite{mstcn} is trained in a supervised way, using the annotated videos, to predict the labels (Fig. \ref{fig:inference_pipeline}). 

Note that our use case requires a multi-label approach. E.g., surgical tools may be used in any of the colon segments, hence the corresponding labels are non-mutually exclusive. Common architectures \cite{mstcn,asformer} used for video segmentation do not support multiple labels natively, and we explain below how to adapt them. 

\subsection{Training Single-Frame Encoder with Key Frames}\label{keyframes}
To improve over the baseline, we follow \cite{lap_chole}, where the frame encoder is pre-trained to predict the labels on a single frame. We use a shared CNN backbone, with multiple classification heads per each class (see Fig. \ref{fig:multihead}). We sample random frames from labeled (annotated) segments, and train the model to predict the segment label. After the training is completed, we discard the classification heads, and use the shared backbone as the feature extractor for the temporal model, as before (Fig. \ref{fig:inference_pipeline}).

\begin{figure}
    \begin{center}
        \begin{tabular}{cc}
            \includegraphics[width=0.2\textwidth]{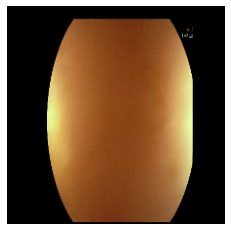} & 
            \includegraphics[width=0.2\textwidth]{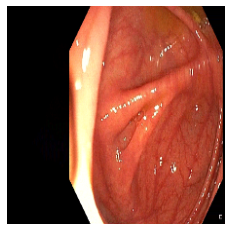} 
        \end{tabular}
    \end{center}
    \caption{Left: A non-informative frame with a blocked field of view. Right: A key frame with a clear view of the triradiate fold.}
    \label{fig:key_frames}
\end{figure}

Interestingly, with this approach, for most labels, we saw performance degradation (Table \ref{backbone_method_ablation_table} - rows 1 vs. 2). We suspect the reason for this is that in colonoscopy many frames are not informative, e.g. due to camera motion blur, liquids, blocked view, etc. Such frames can be frequently found in annotated video segments. Hence, when training on frames randomly sampled from those segments, many of them are not indicative of the segment label (see Fig. \ref{fig:key_frames}). 

Instead, we suggest training the per-frame model with "high-quality" frames that clearly represent the relevant label (Fig. \ref{fig:key_frames}). For example, for the cecum we leverage snapshots manually captured by  gastroenterologist during the procedure to be included in the procedure report. Usually, in those snapshots the cecum landmarks (appendiceal orifice, triradiate fold and ileocecal valve) are clearly visible, hence providing a strong visual signal. We used the snapshots as positives frames to train the cecum head, and took random frames outside of the cecum segment as negatives.
We see a significant improvement (Table \ref{backbone_method_ablation_table} - row 3) of about $4\%$ per-frame accuracy with this approach.  

\subsection{Pseudo Labels}\label{pseudo}

Annotating large video datasets is both time and labor intensive. We can leverage a large pool of unlabeled data in order to increase the amount and diversity of the training samples. This might be important, as, for example, the "outside" (of body) class could potentially contain a wide range of diverse scenes, so it's essential to build a robust model for this task. In this section we focus on improving the single-frame encoder. The incorporation of temporal models follows in the next section. 

Let  $\mathcal{U} = \{(\mathbf{x}^{(1)},\mathbf{y}^{(1)}),\dots, (\mathbf{x}^{(n)},\mathbf{y}^{(n)})\}$ be a small set of available labeled video \textbf{frames}, where $\mathbf{x}^{(i)}$ is the frame and $\mathbf{y}^{(i)}$ is its multi-label annotation. The labeled frames set $\mathcal{U}$ is composed of frames taken from the annotated video segments and snapshots of cecum landmarks and retroflexions, manually captured during the procedure (as explained in the Section~\ref{keyframes}).

Let $K$ be the number of non-mutually exclusive labels, that is,  $\mathbf{y}^{(i)} = (\mathbf{y}_k^{(i)}|_{k=1}^K)$, where $\mathbf{y}_k^{(i)}$ is a one-hot vector for the $k^{th}$ label of the $i^{th}$ sample. In our case, the one-hot vectors correspond to the following non-exclusive labels: tools/no-tool, ileum/cecum/u-turn/other, and inside/outside.

Let $\mathcal{P} = \{\mathbf{v}_1,\dots, \mathbf{v}_m\}$ be a large pool of available unlabeled \textbf{videos}. Let us denote the number of frames in video $\mathbf{v} \in \mathcal{P}$ by $T$.

\subsubsection{Initial Supervised Model}
\begin{figure}[t]
    \centering
    \includegraphics[width=0.5\textwidth]{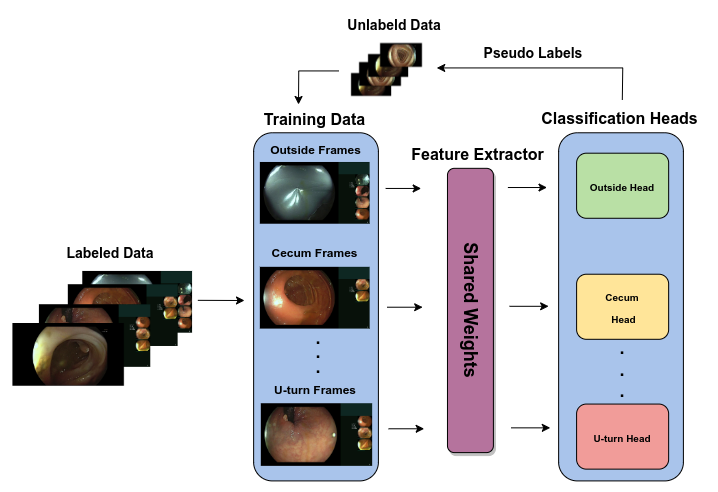} 
    \caption{Pre-training of the feature extractor. We use a combination of labeled data, together with pseudo-labels as explained in Section~\ref{pseudo}. After the training is complete, we discard the classification heads and use the feature extractor to embed frames for the temporal network.}
    \label{fig:multihead}
\end{figure}

We start by training a supervised model $\mathcal{F}$ using the annotated samples from 
$\mathcal{U}$. The model is composed of a shared feature extractor, followed by $K$ different classification heads for each one of the $K$ non-mutually exclusive labels (see Figure \ref{fig:multihead}). We train the model to optimize the cross-entropy loss between the $\mathbf{y}_k^{(i)}$ and the model predictions $ \widehat{\mathbf{y}}_k^{(i)}$:
\begin{equation*}
    L = \sum_{i} \sum_{k} CE(\mathbf{y}_k^{(i)}, \widehat{\mathbf{y}}_k^{(i)}) 
\end{equation*}

In principle, we don't require each sample in $\mathcal{U}$ to have annotations for all $K$ labels. If annotations for some labels are missing, we simply skip them in the summation over $k$.

\subsubsection{Pseudo Labeling and Temporal Smoothing}

To enrich the training set, we label the frames of unlabeled videos in $\mathcal{P}$ by applying the model $\mathcal{F}$ trained using the annotated data, as described in the previous section. 

For every video $\mathbf{v} \in \mathcal{P}$, let $(\mathbf{x}^{(1)}, \dots, \mathbf{x}^{(T)})$ be the $T$ frames of $\mathbf{v}$.  
Let $\widehat{\mathbf{y}}^{(t)}$ be the vector of predicted class probabilities for  frame $\mathbf{x}^{(t)}$:
$$
    \widehat{\mathbf{y}}^{(t)} = \mathcal{F}(\mathbf{x}^{(t)}), t=1..T
$$

As before, $\widehat{\mathbf{y}}^{(t)} = (\widehat{\mathbf{y}}_k^{(t)}|_{k=1}^K)$, where $\widehat{\mathbf{y}}_k^{(t)}$ is the vector of class probabilities for label $k$. In practice, as some of the tasks are difficult to predict from a single frame (cecum detection for example), we use the pseudo labeling approach only for inside-body/outside-body and tools detection tasks.

In order to reduce the pseudo-label noise, we smooth the class predictions along the temporal dimension $t$ by a Gaussian kernel $G_\sigma$ of size $2M+1$ to yield
$$
\widetilde{\mathbf{y}}^{(t)}_k = (\widehat{\mathbf{y}}_k \ast G_\sigma)^{(t)} = \sum_{m=-M}^{M} \widehat{\mathbf{y}}_k^{(t-m)} G_\sigma[m]
$$

\subsubsection{Temporal Consistency Filtering}

In order to further improve the quality of the pseudo-labeling, we leverage the prior domain knowledge about the temporal structure of colonoscopy procedures. We know that a procedure usually (but not always) follows the predefined sequence of phases:

\begin{enumerate}
    \setlength\itemsep{0.05em}
    \item Outside of the body
    \item Inside the body
    \item Outside of the body
\end{enumerate}

We choose a very simple sanity check approach to discard videos with pseudo-labels that do not follow the very basic outside-inside-outside temporal pattern:
Let us denote by $\widetilde{\mathbf{y}}^{(t)}_\text{in/out}$ the predicted inside/outside label class probability vector for frame $t$, where $in/out$ is the index of the corresponding label.
We require the start and end frames of the video to be outside of the body, and the middle frame of the video to be inside: 
\begin{eqnarray}
\argmax_{l\in(0,1)} \widetilde{\mathbf{y}}^{(0)}_\text{in/out}[l] = \argmax_{l\in(0,1)}  \widetilde{\mathbf{y}}^{(T)}_\text{in/out}[l] = 1, \nonumber \\ \nonumber
\argmax_{l\in(0,1)} \widetilde{\mathbf{y}}^{(T/2)}_\text{in/out}[l] = 0. \nonumber
\end{eqnarray}
In addition, tools almost never appear outside the body, as they are usually visible once the endoscope is inside the body and the physician is examining a polyp.
Hence, videos in which we detect tools and outside the body over the same frame are also discarded.
The reason we limit the temporal filtering to these simple heuristics is because complex temporal dependencies are introduced through a temporal network, as described in the following section. At this stage we are only interested to make sure the generated pseudo-labels are of reasonable quality, to reduce the label noise while training the single-frame encoder. 

After discarding videos that don't meet these criteria, we re-train $\mathcal{F}$ to predict $\widetilde{\mathbf{y}}^{(i)}$ for pseudo-labeled videos $\mathbf{v} \in \mathcal{P}$ (for inside/outside and tools detection tasks), in addition to the annotated samples in $\mathcal{U}$. This way we significantly increase the size and the diversity of the training set.

\begin{figure*}[t]
    \centering
    \includegraphics[width=0.8\textwidth]{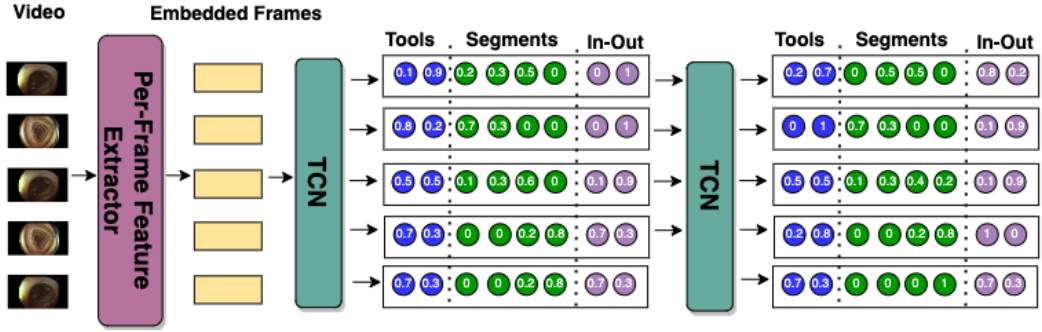} 
    \caption{Multi-Label MS-TCN with two stages (the number of stages is a hyperparameter). Note that we apply the Softmax activation separately on the logits that correspond to the colon-segments, inside/outside and tools/no-tools.}
\label{fig:multiehad_mstcn}
\end{figure*}

\begin{table*}
\centering
\begin{tabular}{|l|l|l|l|l|l|l|l|}
\hline
\scriptsize\bfseries Architectures & \scriptsize\bfseries Avg. Accuracy &\scriptsize\bfseries Ileum &\scriptsize\bfseries Cecum & \scriptsize\bfseries \makecell{Rectal \\ Retroflexion}  & \scriptsize\bfseries Outside & \scriptsize\bfseries Tool \\
\hline
\scriptsize\bfseries  ResNet, MS-TCN & $90.4\pm0.8$ & $90.5\pm0.7$ & $89.6\pm0.8$ & $96.3\pm1.0$ & $\textbf{99.8}\pm0.1$ & $\textbf{93.7}\pm0.9$\\
\hline
\scriptsize\bfseries  ResNet, ASFormer & $90.4\pm0.9$ & $88.8\pm1.8$ & $90.7\pm0.4$ & $97.1\pm0.7$ & $99.7\pm0.2$ & $91.0\pm1.3$\\
\hline
\scriptsize\bfseries  ConvNext, MS-TCN & $94.1\pm0.5$ & $94.4\pm0.9$ & $\textbf{92.5}\pm0.3$ & $98.7\pm0.5$ & $\textbf{99.8}\pm0.1$ & $\textbf{93.7}\pm0.4$\\
\hline
\scriptsize\bfseries  ConvNext, ASFormer & $\textbf{94.6}\pm0.5$ & $\textbf{96.1}\pm0.5$ & $91.5\pm0.7$ & $\textbf{99.0}\pm0.3$ & $\textbf{99.8}\pm0.1$ & $92.1\pm2.0$\\
\hline
\end{tabular}
\caption{Ablation study for network architectures. Average per-frame balanced classification accuracy over all labels, and for each label. Models trained with the "key-frame" training scheme. }
\label{architectures_ablation_table}
\end{table*}

\begin{table*}
\centering
\begin{tabular}{|l|l|l|l|l|l|l|}
\hline
\scriptsize\bfseries Method & \scriptsize\bfseries Avg. Accuracy &\scriptsize\bfseries Ileum &\scriptsize\bfseries Cecum & \scriptsize\bfseries \makecell{Rectal \\ Retroflexion}  & \scriptsize\bfseries Outside & \scriptsize\bfseries Tool \\
\hline
\scriptsize\bfseries  ImageNet pre-training & $90.8\pm1.0$ & $92.3\pm2.3$ & $90.3\pm0.8$ & $94.9\pm2.0$ & $99.6\pm0.3$ & $90.2\pm2.4$\\
\hline
\scriptsize\bfseries  Classification on random labeled frames & $87.2\pm1.4$ & $88.8\pm2.7$ & $89.5\pm0.8$ & $90.4\pm3.5$ & $99.7\pm0.0$ & $86.9\pm2.4$\\
\hline
\scriptsize\bfseries  Classification on key frames & $\textbf{94.6}\pm0.5$ & $\textbf{96.1}\pm0.5$ & $\textbf{91.5}\pm0.7$ & $\textbf{99.0}\pm0.3$ & $\textbf{99.8}\pm0.1$ & $\textbf{92.1}\pm2.0$\\
\hline
\end{tabular}
\caption{Ablation study for frame encoder training scheme. Average per-frame balanced classification accuracy over all labels, and for each label. Using ConvNextBase as the frame encoder and ASFormer as the temporal model.}
\label{backbone_method_ablation_table}
\end{table*}

\subsection{Multi-Label Temporal Network}

The main design improvement of MS-TCN~\cite{mstcn} over the TCN~\cite{TCN} architecture, is the multi-stage approach. The first stage takes the frame embeddings and predicts a class for each frame (as in TCN), while the following stages "refine" those predictions. That is, the next stages are fed with the class-predictions of the previous stage, and the predictions of all stages equally contribute to the loss. A multi-stage design is also used by more modern action segmentation networks, such as the transformer-based ASFormer~\cite{asformer}. We cannot apply the MS-TCN approach in our case, as it does not support multi-label classification. To the best of our knowledge, there is no prior art that applies a multi-stage temporal network for a multi-label problem. 

A naive adaptation of MS-TCN to deal with multi-label is to use separate networks for each label, at least from the 2nd stage on. This problem with this approach is that it does not allow any cross-talk between network signals corresponding to different labels. Our design allows all stages to exchange information related to different labels. This might be beneficial, as, for example, it is less common to see a tool in the ileum or outside of the body, and we would like the model to learn these priors and use them to refine predictions in later stages of the network. To enable this, we feed each stage with the concatenation of all class probabilities for all labels. 

More formally, let $(\mathbf{x}^{(1)}, \dots \mathbf{x}^{(T)})$ be the $T$ frames of a video. Each frame $\mathbf{x}^{(t)}$ is labeled with $K$ different labels $(\mathbf{y}^{(t)}_1, \dots, \mathbf{y}^{(t)}_K$). In our case $K=3$, and $(\mathbf{y}^{(t)}_1, \mathbf{y}^{(t)}_2, \mathbf{y}^{(t)}_3)$ are the 2-,4- and 2-long one-hot vectors,  corresponding to tool/no-tool, ileum /cecum/rectal-retroflextion/other, and inside/outside labels respectively.
Let $\widehat{\mathbf{y}}^{(t)}_k$ be the vector of predicted class probabilities for $k^{th}$ label of frame $\mathbf{x}^{(t)}$. 

Our solution for the multi-label setup uses the per-label softmax applied to groups of logits corresponding to each label (see Fig. \ref{fig:multiehad_mstcn}). The concatenated  probabilities vector is then fed into the next stage.

Assume the network has $S$ stages.
Denote by $O^{(t)}_s$ the output of the $s^{th}$ stage for the frame $\mathbf{x}^{(t)}$ (pre-softmax), and by $I_{k}$ the indices of the logits relevant to the $k^{th}$ label. Then the predicted vector of class probabilities for the $k^{th}$ label, ${s^{th}}$ stage, and $t^{th}$ frame is
\begin{equation}
\widehat{\mathbf{y}}^{(t)}_{k,s} = \textbf{softmax}(O^{(t)}_s[j]|{j \in I_{k}}).
\end{equation}
The corresponding loss term, as defined in MS-TCN \cite{mstcn} is:
\begin{equation}
l^{(t)}_{k,s} = \frac{1}{T} CE(\widehat{\mathbf{y}}^{(t)}_{k,s}, \mathbf{y}^{(t)}_k) + \lambda \frac{1}{T |I_{k}|}|| \widehat{\mathbf{y}}^{(t-1)}_{k,s} - \widehat{\mathbf{y}}^{(t)}_{k,s}||^{2}_2,
\label{eq:loss}
\end{equation}
where $CE$ is the cross entropy loss between the ground truth for the $t^{th}$ frame and the $k^{th}$ label $\mathbf{y}^{(t)}_{k}$ and the prediction $\widehat{\mathbf{y}}^{(t)}_{k,s}$. The second loss term is a smoothing loss that encourages adjacent frames to have similar predictions. $\lambda$ is a weighting factor.
The final loss is computed over all stages, frames and labels:
\begin{equation}
loss = \sum_{s=0}^{S}\sum_{t=0}^{T}\sum_{k=0}^{K} w_{k} \cdot l^{(t)}_{k,s}, 
\label{eq:warp_expand}
\end{equation}
where $w_{k}$ is a per-label weighting factor.

The proposed scheme enables multiple labels per frame, while introducing minimal changes to the original MS-TCN architecture.

\section{Experiments and Results}\label{experiments}
\subsection{Dataset}
Our labeled data consists of 3,994 colonoscopy videos, recorded in 3 medical centers. We randomly split it into 344 videos for testing and 3,650 for training. We do not use a validation dataset as we do not perform hyper-paramter tuning in this paper, and leave this for future work. Instead, we use the commonly used parameters as described in Tables \ref{per_frame_params_table} and \ref{temporal_params_table}.
We make use of labeled video segments annotated by experienced gastroenterologists (see Table \ref{annotation_table}). 
For each video, time segments were labeled, indicating when different colon-segments/tools appear, or whether the endoscope is outside of the body. 
For key-frames we use still-images of anatomical landmarks, which gastroenterologists manually captured during the colonoscopy procedure. In addition, we leverage the unlabeled set of 18,500  colonoscopy videos, by training the model on pseudo labels computed for these videos, as explained in Section~\ref{pseudo}.

All videos were standardized to 30 FPS, and had original resolution of 720P or 1080P (later we down-sample to $224 \times 224$ for training). The minimum bitrate used for compression was 12mbps, and the median procedure time is 11 minutes. Finally, the videos were captured by multiple endoscope types from 3 different manufactures: Olympus, Fuji and Pentax.

\begin{table}[h]
\centering
\begin{tabular}{|l|l|l|}
\hline
\scriptsize\bfseries  Activity / Segment & \scriptsize\bfseries  Snapshots & \scriptsize\bfseries  Annotated Segments \\
\hline
\scriptsize\bfseries Cecum &  65K & 2.5K\\
\hline
\scriptsize\bfseries Rectal Retroflexion & 20K & 1.5K\\
\hline
\scriptsize\bfseries Tools & - & 14K\\
\hline
\scriptsize\bfseries Terminal Ileum & - & 1K\\
\hline
\scriptsize\bfseries Inside and Outside & - & 1.5K\\
\hline
\end{tabular}
\caption{Annotations: number of annotated segments and stills for each label.}
\label{annotation_table}
\end{table}

\subsection{Accuracy Evaluation and Ablation Study}
We preform an ablation study to better understand the role of different components. In particular, we compare the single frame encoder pre-trained on the Imagenet, with the one trained on random frames from annotated segments, and the one trained on key-frames.
We measure the average per-frame classification accuracy on the test set, over all labels, and for each label. For the per-label results we use balanced accuracy (with equal weights for sensitivity and specificity) as the labels are heavily unbalanced. For each setup we ran 5 experiments and report the average and standard deviation. As can be seen in Table~\ref{backbone_method_ablation_table}, our method with the key-frames training significantly outperforms the random sampling and the ImageNet baseline.
As explained in Section~\ref{keyframes}, for most labels (ileum, cecum, rectal retroflextion and tools), the random sampling actually hurts the performance.   

\begin{table}[h]

\centering
\begin{tabular}{|l|l|l|}
\hline
\scriptsize\bfseries Parameter & \scriptsize\bfseries ResNet50v2 & \scriptsize\bfseries ConvNextBase \\
\hline
\scriptsize\bfseries{Batch Size} & 64 & 64 \\
\hline
\scriptsize\bfseries{Optimizer} & Adam & Adam \\
\hline
\scriptsize\bfseries{Hardware} & 4 Tesla v100 & 4 Tesla v100 \\
\hline
\scriptsize\bfseries{Num. of Param.} & 24M & 88M \\
\hline
\scriptsize\bfseries{Gaussian kernel} & $\sigma = 5,  M = 10$ & $\sigma = 5, M = 10$ \\
\hline
\scriptsize\bfseries{Resolution} & $224 \times 224$ & $224 \times 224$ \\
\hline
\end{tabular}
\caption{Per-Frame Embedding Model training setup parameters.}
\label{per_frame_params_table}
\end{table}

\begin{table}[h]
\centering
\begin{tabular}{|l|l|l|}
\hline
\scriptsize\bfseries Parameter & \scriptsize\bfseries{MS-TCN} & \scriptsize\bfseries ASFormer \\
\hline
\scriptsize\bfseries{Batch Size} & 1 & 1 \\
\hline
\scriptsize\bfseries{Optimizer} & Adam & Adam \\
\hline
\scriptsize\bfseries{Stages} & 2 & 2 \\
\hline
\scriptsize\bfseries{Layers per Stage} & 13 & 9 \\
\hline
\scriptsize\bfseries{$\lambda$ Smoothing loss factor} & 0.15 & 0.15 \\
\hline
\scriptsize\bfseries{Hardware} & 1 Tesla v100 & 1 Tesla v100 \\
\hline
\scriptsize\bfseries{Num. of Param.} & 0.5M & 0.6M \\
\hline
\end{tabular}
\caption{Temporal Model training setup parameters.}
\label{temporal_params_table}
\end{table}

We also compare several network architectures: ResNet50~\cite{resnet} and ConvNextBase~\cite{convnext} for feature extractors, and MS-TCN~\cite{mstcn} and ASFromer~\cite{asformer} for temporal networks (Table \ref{architectures_ablation_table}). 
We notice a significant improvement with the larger and more modern ConvNextBase compared to ResNet50. On the other hand, the results of ASFromer and MS-TCN seem on-par.
For the training settings and hyperparameters see Tables \ref{per_frame_params_table} and \ref{temporal_params_table}.
Overall, as one can see, our proposed method achieves very high accuracy, reaching high 90s for most labels.

\section{Conclusions and Future Work}
We presented a method for semantic parsing of colonoscopy videos. The proposed technique adapts the multi-stage temporal network (MS-TCN) to a multi-label scenario. To gain more accuracy, we improve the single frame feature extractor by training it on key-frames and pseudo-labeling. The method is evaluated on hundreds of colonoscopies and demonstrates above $90\%$ accuracy for all labels. Semantic parsing of colonoscopy videos enables a number of downstream applications, including quality metrics, video retrieval, and automatic report generation. 

There are several promising directions for future work that can be based on this method and expand its capabilities. We plan adding additional colon segments such as the transverse, ascending, and descending colon. Another direction is automatic detection of various colonoscopy imaging modes including Narrow Band Imaging (NBI) and chromoendoscopy. Pursuing these avenues introduces more automation to colonoscopy, contributing to more accurate and efficient diagnosis and treatment.

{\small
\bibliographystyle{ieee_fullname}
\bibliography{egbib}
}

\end{document}